\documentclass[lettersize,journal]{IEEEtran}

\usepackage{times}
\usepackage{epsfig}
\usepackage{graphicx}
\usepackage{amsmath}
\usepackage{amssymb}
\RequirePackage[format=plain,labelformat=simple,labelsep=period,font=small,compatibility=false]{caption}
\RequirePackage[font=footnotesize,skip=3pt,subrefformat=parens]{subcaption}

\usepackage{booktabs}
\usepackage{tikz}
\usetikzlibrary{matrix,fit}
\usepackage{calc}
\usepackage{numprint}
\usepackage{siunitx}
\usepackage{float}
\usepackage{cite}
\usepackage{xspace}
\usepackage{dblfloatfix}

\usepackage{etoolbox}
\newrobustcmd\B{\DeclareFontSeriesDefault[rm]{bf}{b}\bfseries} 

\usepackage[pagebackref=true,breaklinks=true,bookmarks=false]{hyperref}

\usepackage[capitalize]{cleveref}
\crefname{section}{Sec.}{Secs.}
\Crefname{section}{Section}{Sections}
\Crefname{table}{Table}{Tables}
\crefname{table}{Tab.}{Tabs.}

\newcommand{\ve}[1]{\mathbf{#1}}
\newcommand{\nv}{\ve{n}}

\newcommand{\vv}{\ve{v}}

\newcommand{\vpv}{\ve{v'}}
\newcommand{\dotp}[2]{#1^T #2}
\newcommand{\comment}[1]{}

\newcommand*\matr{\mathbf}
\newcommand{\Aff}{\matr{A}}

\newcommand{\Pl}{\matr{\Pi_1}}
\renewcommand{\Pr}{\matr{\Pi_2}}
\newcommand{\Plx}{\ve{\Pi_{1u}}}
\newcommand{\Ply}{\ve{\Pi_{1v}}}
\newcommand{\Prx}{\ve{\Pi_{2u}}}
\newcommand{\Pry}{\ve{\Pi_{2v}}}

\newcommand{\Plu}{\ve{\Pi_{1u}}}
\newcommand{\Plv}{\ve{\Pi_{1v}}}
\newcommand{\Pru}{\ve{\Pi_{2u}}}
\newcommand{\Prv}{\ve{\Pi_{2v}}}

\DeclareRobustCommand\onedot{\futurelet\@let@token\@onedot}
\def\@onedot{\ifx\@let@token.\else.\null\fi\xspace}
\def\etal{\emph{et al}\@onedot}

\begin{document}

\title{Robust and Real-time Surface Normal Estimation from Stereo Disparities using Affine Transformations}

\author{Csongor Csanád Karikó, Muhammad Rafi Faisal, and Levente Hajder\\
Geometric Computer Vision Group, Faculty of Informatics\\
Eötvös Loránd University, Budapest, Hungary\\
{\tt\small \{jpoiwr,goo2im,hajder\}@inf.elte.hu}\\
}

\maketitle

 \begin{abstract}
This work introduces a novel method for surface normal estimation from rectified stereo image pairs, leveraging affine transformations derived from disparity values to achieve fast and accurate results. We demonstrate how the rectification of stereo image pairs simplifies the process of surface normal estimation by reducing computational complexity. To address noise reduction, we develop a custom algorithm inspired by convolutional operations, tailored to process disparity data efficiently. We also introduce adaptive heuristic techniques for efficiently detecting connected surface components within the images, further improving the robustness of the method. By integrating these methods, we construct a surface normal estimator that is both fast and accurate, producing a dense, oriented point cloud as the final output. Our method is validated using both simulated environments and real-world stereo images from the Middlebury and Cityscapes datasets, demonstrating significant improvements in real-time performance and accuracy when implemented on a GPU. Upon acceptance, the shader source code will be made publicly available to facilitate further research and reproducibility.
\end{abstract}

\section{Introduction}
\label{sec:intro}

\begin{figure}
    \centering
    \includegraphics[width=\linewidth]{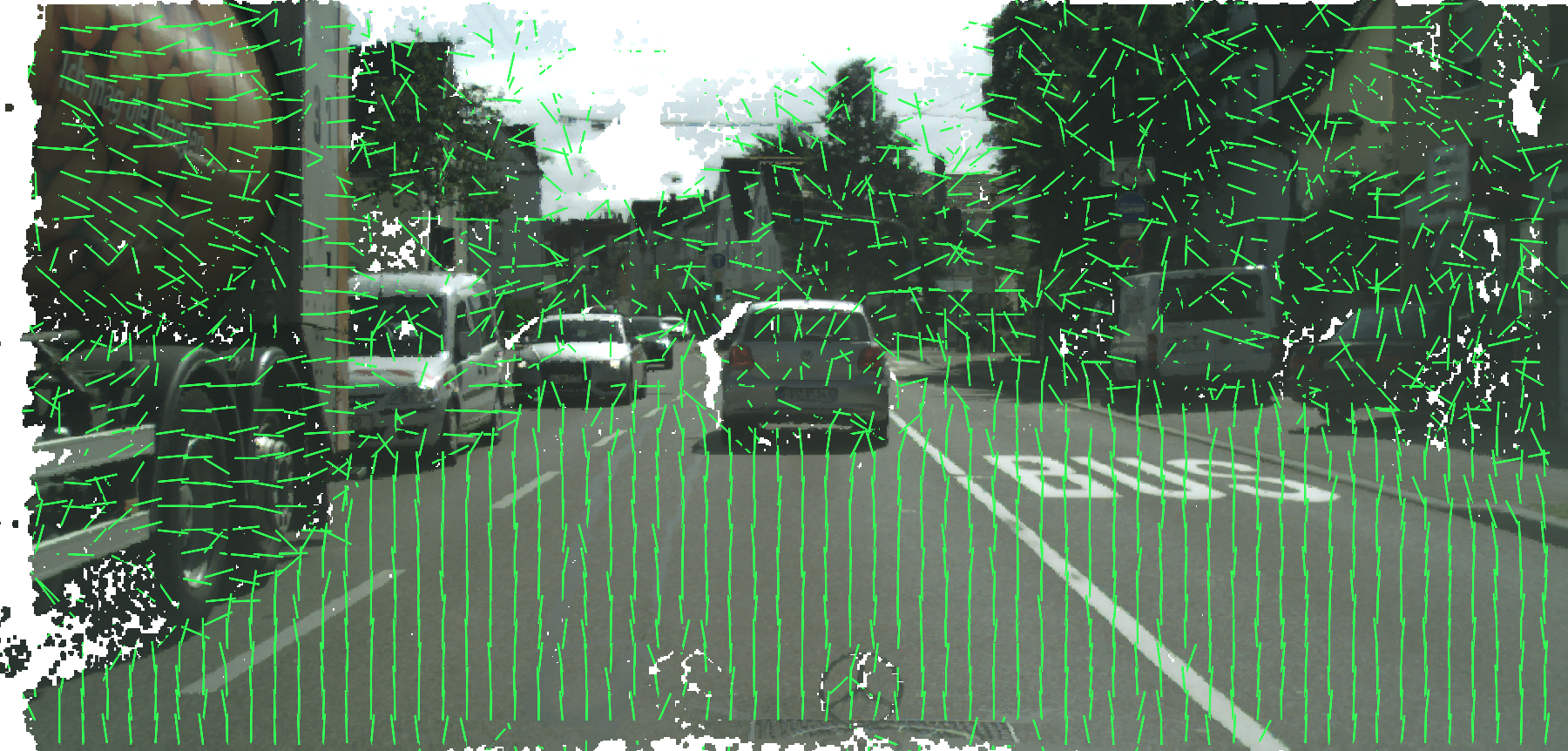}
    \caption{Estimated surface normals are drawn in a vehicle-mounted camera image from the Cityscape dataset~\cite{Cordts2016Cityscapes}. For the road, normals are vertical, reconstruction of those can significantly improve e.g. the segmentation of road from other objects. \textit{Normals are estimated by the proposed approach.}}
    \label{fig:intro_car}
\end{figure}

Obtaining oriented point clouds is a crucial challenge in both computer vision and robotics, with numerous real-world applications reaping the advantages of the additional information they offer.
These applications encompass tasks such as 3D reconstruction\cite{DBLP:journals/cacm/AgarwalFSSCSS11, DBLP:conf/cvpr/BarathMESM21, DBLP:conf/cvpr/HeinlySDF15, DBLP:conf/cvpr/SchonbergerF16, DBLP:conf/cvpr/ZhuZZSFTQ18, DBLP:journals/ijcv/SnavelySS08, Raposo2020} and simultaneous localization and mapping\cite{DBLP:journals/corr/DeToneMR17, DBLP:conf/eccv/EngelSC14, DBLP:journals/trob/Mur-ArtalMT15} (SLAM), where the orientation of 3D points offers valuable insights into the underlying structure. See \cref{fig:intro_car} for an example. Image-based visual localization methods\cite{DBLP:journals/ijrr/LynenZABHPSS20, DBLP:conf/eccv/PanekKS22, DBLP:conf/eccv/SattlerLK12, DBLP:conf/cvpr/SattlerMTTHSSOP18} can also leverage knowledge of surface normals. In the context of autonomous vehicles, understanding surface orientation can enhance tasks like ground or facade segmentation within visual odometry\cite{DBLP:conf/cvpr/NisterNB04, DBLP:journals/jfr/NisterNB06}. Furthermore, the use of oriented point clouds greatly simplifies geometric model and multi-model estimation~\cite{isack2012energy,barath2018multi,barath2019progressive} by enabling the development of solvers that require fewer data points than their point-based counterparts. Additionally, surface normals play a significant role in the numerical refinement of the reconstructed oriented 3D point cloud~\cite{Hajder2017}.

Nowadays, 3D reconstruction pipelines usually serve the result as 3D point clouds, and the surface orientation is computed from spatial points.
This paper principally addresses the rapid and accurate estimation of surface normals based on affine correspondences.

\subsection{Surface Normal Estimation}
There are several different techniques to estimate surface normals from stereo images, they are overviewed here.

\noindent \textit{Surface normal from point clouds:} Upon applying state-of-the-art sparse\cite{schonberger2016structure} or dense~\cite{DBLP:conf/nips/KarHM17,chen2019point} reconstruction pipelines, the resultant point cloud can be readily endowed with point orientations. In this scenario, it is common practice to estimate normals on a per-point basis, employing the widely accepted Principal Component Analysis (PCA)~\cite{Jolliffe2016} to deduce tangent planes  from the immediate local vicinity surrounding each point. Higher-order surfaces~\cite{Benko2002,Jiao2012} can also be reconstructed from point clouds. Nevertheless, this approach has two significant drawbacks. First, the choice of neighborhood size significantly impacts accuracy and is often uncertain in the absence of a metric reconstruction. Alternatively, using the k-nearest-neighbors algorithm to determine local neighborhoods is a feasible solution, but the estimation is heavily influenced by the the sparsity of the point cloud. Second, independently fitting local surfaces to the neighborhood of each 3D point in the reconstruction is an exceedingly resource-intensive operation.

\noindent \textit{From affine transformations:}
Recently, the mainstream point-based reconstruction systems have been extended by the application of affine transformations~\cite{Hajder2017}. An affine transformation connects two small corresponding regions between two images. They are the linear local approximation of differentiable $\mathbb{R}^2\longrightarrow\mathbb{R}^2$ functions. The pioneering work in this area was the PhD thesis of Kevin Koser~\cite{Koser2008} in 2008. Bentolilla and Francos~\cite{bentolila2014conic} showed that the epipoles can be efficiently estimated via affine transformation. Molnar and Chetverikov introduced a general formula~\cite{Molnar2014} that shows how surface normals, projected coordinates, camera parameters determine the related affine transformation. Later on, Barath \etal~\cite{BarathH17} showed how homographies can be estimated from fewer correspondences if affine parameters are also available. Remark that the normal can be retrieved from a homography~\cite{Raposo2020} if the camera intrinsic parameters are known.  The essential~\cite{Raposo016} and fundamental~\cite{barath2016novel} matrices can also be estimated from two and three correspondences, respectively.

The first normal estimator for a general pin-hole camera pair was proposed by Barath \etal~\cite{barath2016novel} in 2015. Since their work, faster methods~\cite{Minh20,Hajder2023ICCV} have appeared in the literature. 

\noindent \textit{Surface normals via machine learning:} Nowadays, machine learning has become a frequently used method in the computer vision community.  There are specific networks~\cite{EigenF15,BansalRG16,ZhanWGR19,WangFG15} for normal estimation as well. However, their results highly depend on the training data, and one of our main purposes is to develop a deterministic approach as surface normal can be computed from the disparities.

\subsection{Goals}
Although there are general estimators~\cite{barath2016novel,Minh20,Hajder2023ICCV} to compute the surface normals from an  affine correspondence if the cameras are calibrated, there are many open problems:
\begin{itemize}
    \item The \textit{speed is crucial} especially \textit{in real-time applications} if a very dense oriented point cloud is required as a result.
    \item The accuracy of surface normal estimation is \textit{very sensitive} to the quality of the affine transformations, sophisticated filtering methods have not been proposed before, to the best of our knowledge. 
    \item Affine transformations are the first-order approximations of the projection between corresponding image patches. However, theoretically, the approximation is valid only if the pixels are related to the same surfaces. At their borders, multiple surfaces might contribute pixels to the estimation of the same normal vector. A \textit{major} task is therefore to detect the borders of these surfaces.
\end{itemize}
We address all these three listed goals in our paper.


\subsection{Contribution}
 
The main theoretical novelty here is that rectified images are applied for affine transformation estimation for the sake of lower processing time. This special stereo image setup can be achieved from a general image pair if the epipoles are not on the image. There are efficient methods to rectify image pairs~\cite{Hartley1999}. Therefore, it is not a strict restriction. 

To the best of our knowledge, only Megyesi \etal~\cite{Megyesi2006} dealt with the application of affine transformations for rectified images. Their method is not complete as they only proposed a simple formula to determine the normal from the affine transformation and that formula does not consider the camera intrinsic parameters. Their work highly motivated us, but we replace their formula with an alternative one, including camera intrinsics, and we also deal with the accurate estimation of affine transformations and border detection of planar regions.

As speed is also a critical aspect, the application of GPUs is preferred. Modern GPU architectures are massively parallel and have a certain set of limitations that need to be taken into account when designing and implementing algorithms for them. The SIMD architecture employed by modern GPUs executes the same instruction in parallel over multiple sets of data. Therefore usually it is costly to do a lot of branching if the code is expected to be executed differently on neighboring pixels. It is also generally advisable to use as little extra memory as possible per pixel. If the required memory is too much to be entirely stored in local registers, performance degrades significantly.
We designed our algorithms considering these limitations.

The key \textbf{contribution} of our work \textbf{is fourfold}:  \textbf{(i)} It is shown how the normal estimation is simplified for a rectified image stereo pair. \textbf{(ii)} We propose a novel method here that computes the normal vector from affine transformations determined between rectified images. \textbf{(iii)} A novel algorithm is also introduced to retrieve the affine parameters from disparity values. It works for both the minimal and over-determined cases. \textbf{(iv)} An adaptive method for border detection of regions corresponding to flat surfaces is proposed.

Our approach is purely geometric, without any machine learning components. Therefore, learning data is not required, the results are \textbf{deterministic}, it depends only on the disparities. Nevertheless, the proposed methods are straightforwardly differentiable, with the exception of the adaptive estimation, thus the approach can be built into a complex system as a component for which end-to-end training is possible. Moreover, our geometry-based solution contains a convolution, this fact suggests that there are similarities between the operations of Convolutional Neural Networks (CNNs) and our geometric approach.


\section{Theoretical Background}

The relationship between inter-region affine transformations of stereo images and the surface normal has been studied previously by Barath \etal~\cite{barath2016novel}. Here, we deduce the related formulas for a rectified stereo pair.

In general, the affine transformation is derived~\cite{barath2016novel} from the surface normal as follows:
\begin{equation}
\label{eq:affine-matrix-general}
    \Aff = \begin{bmatrix}
    a_1 & a_2\\
    a_3 & a_4
    \end{bmatrix} =
    \frac{1}{\nv^T \ve{w_5}}\begin{bmatrix}
        \nv^T\ve{w_1} & \nv^T\ve{w_2}\\
        \nv^T\ve{w_3} & \nv^T\ve{w_4}
    \end{bmatrix},
\end{equation}
where $\nv$ denotes the surface normal and $\ve{w_1}$ through $\ve{w_5}$ are calculated from the gradients of the projection matrices. These can be written as follows: 
\comment{\begin{align}
    w_1 &= \plv\times\pru
    w_2 &= \pru\times\plu\\
    w_3 &= \plv\times\prv
    w_4 &= \prv\times\plu\\
    w_5 &= \plv\times\plu
\end{align}}
$\ve{w_1}=\nabla\Plv\times\nabla\Pru$, $\ve{w_2}=\nabla\Pru\times\nabla\Plu$, $\ve{w_3}=\nabla\Plv\times\nabla\Prv$, $\ve{w_4}=\nabla\Prv\times\nabla\Plu$ and $\ve{w_5}=\nabla\Plv\times\nabla\Plu$.

In the formulas, $\mathbf \Pi_i(x,y,z)$, $i \in \{1,2\}$, are the projection functions transforming world to image coordinates as 

\begin{equation}
    \mathbf \Pi_i(x,y,z)=\begin{bmatrix}
    \mathbf \Pi_{iu}(x,y,z)\\
    \mathbf \Pi_{iv}(x,y,z)
\end{bmatrix}.
\end{equation}


\begin{figure}
    \centering
    \includegraphics[width=0.7\linewidth]{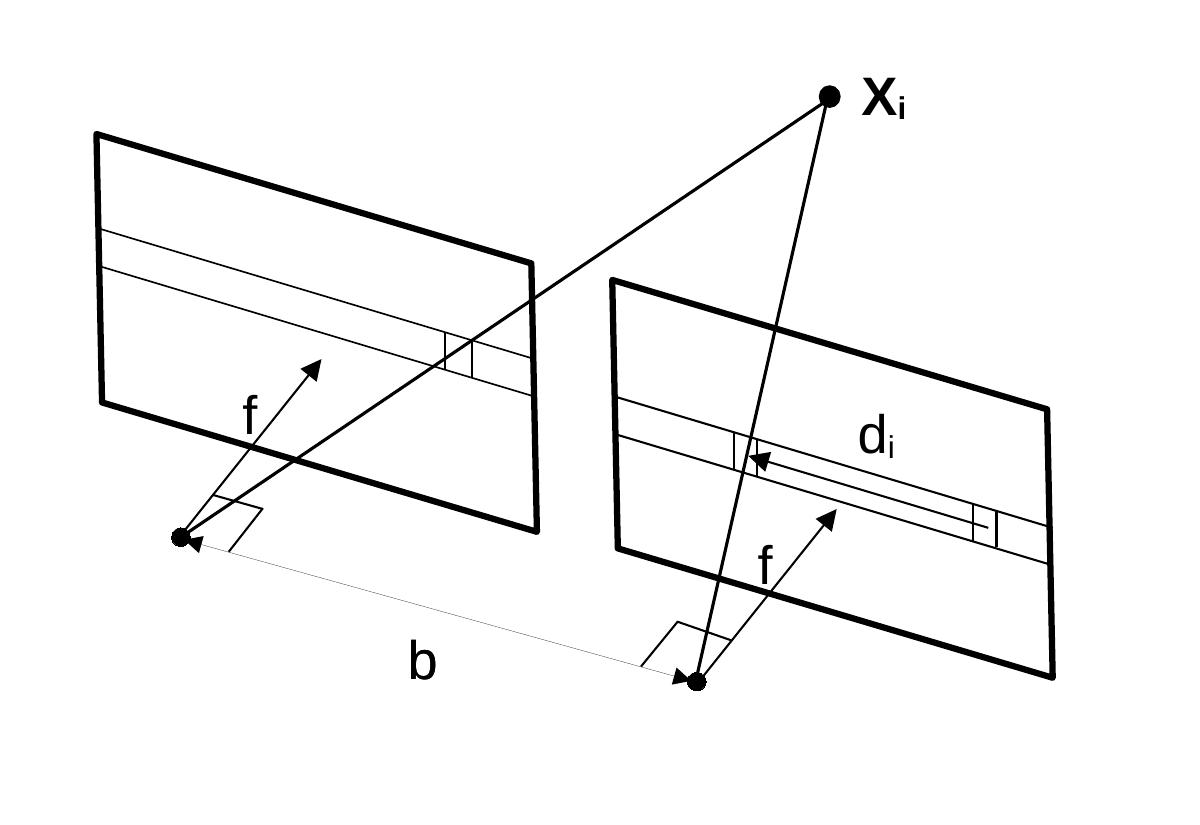}
    \caption{Geometric setup for rectified stereo. It is represented by baseline $b$ and common focal length $f$. The inputs are the disparity values $d_i$ coming from the projection of spatial points $\mathbf X_i$.}
    \label{fig:intro}
\end{figure}

\noindent \textbf{Rectified stereo images.} \Cref{fig:intro} shows the special case of a rectified stereo setup with pin-hole cameras when the intrinsic parameters are the same for both camera. In this case the projection functions can be written as
\begin{eqnarray}
    \Pl(x,y,z)=&\frac{1}{z}\begin{bmatrix}
        f_xx+u_0z\\
        f_yy+v_0z
    \end{bmatrix}, \\
    \Pr(x,y,z)=&\frac{1}{z}\begin{bmatrix}
        f_x(x-b)+u_0z\\
        f_yy+v_0z
    \end{bmatrix},
\end{eqnarray}
where $x,y,z$ are the world coordinates, $f_x$ and $f_y$ are the horizontal and vertical focal lengths in pixels, $b$ is the baseline distance between the cameras and $[u_0,v_0]^T$ is the principal point in which the optical axis intersects the image plane.

The gradients of the projective functions for the pin-hole camera can be written as follows:
\begin{align}
    &\nabla\Plx = \frac{1}{z^2}\begin{bmatrix}
        f_xz & 0 & -f_xx
    \end{bmatrix}^T,\\
    &\nabla\Ply = \frac{1}{z^2}\begin{bmatrix}
        0 & f_yz & -f_yy
    \end{bmatrix}^T,\\
    &\nabla\Prx = \frac{1}{z^2}\begin{bmatrix}
        f_xz & 0 & f_x(b-x)
    \end{bmatrix}^T,\\
    &\nabla\Pry = \nabla\Ply.
\end{align}

\noindent After calculating the cross product of the respective gradients, one can get for $\ve{w_1}\ldots\ve{w_5}$ that
\begin{align}
\label{eq:w1}
    &\ve{w_1} = \frac{f_xf_y}{z^3}\begin{bmatrix}
        b-x & -y & -z
    \end{bmatrix}^T,\\
    \label{eq:w2}
    &\ve{w_2} = \frac{f_x^2b}{z^3}\begin{bmatrix}
        0&1&0
    \end{bmatrix}^T,\\
    \label{eq:w3}
    &\ve{w_3} = \begin{bmatrix}
        0&0&0
    \end{bmatrix}^T,\\
    \label{eq:w45}
    &\ve{w_4} = \ve{w_5} = -\frac{f_xf_y}{z^3}\begin{bmatrix}
        x&y&z
    \end{bmatrix}^T.
\end{align}

\noindent Substituting \eqref{eq:w1}, \eqref{eq:w2}, \eqref{eq:w3},  \eqref{eq:w45} into \eqref{eq:affine-matrix-general} yields
\begin{eqnarray}
    \nonumber
    \Aff =\begin{bmatrix}
        \frac{\nv^T\ve{w_1}}{\nv^T\ve{w_4}} & \frac{\nv^T\ve{w_2}}{\nv^T\ve{w_4}} \\
        \frac{\nv^T\ve{0}}{\nv^T\ve{w_4}} & \frac{\nv^T\ve{w_4}}{\nv^T\ve{w_4}}
    \end{bmatrix} = 
    \begin{bmatrix}
        a_1 & a_2 \\
        0 & 1
    \end{bmatrix}= \\
    \label{eq:affine_mtx}
\label{eq:affine-matrix-expanded}
    \begin{bmatrix}
        \frac{n_x(x-b)+n_yy+n_zz}{n_xx+n_yy+n_zz} & \frac{-bf_xn_y}{f_y(n_xx+n_yy+n_zz)}\\
        0&1
    \end{bmatrix}.
\end{eqnarray}

\section{Proposed Method}
In this section, it is shown how a robust surface normal estimator for a rectified stereo image pair can be formed.  First, it is deduced how the normal can be determined using the affine parameters. Then it is derived how these affine parameters can be computed from disparity values. Finally, an adaptive method is proposed to robustly detect the area from which disparities are selected.

\subsection{Normal Calculation}
If we apply the Fast Normal Estimation (FNE) formulas from \cite{barath2016novel} in order to calculate the normal, this special case always yields a zero vector as $a_3=0$ is a scalar multiplier.
For this reason, we derived a new method for calculating the normal in this special case. 

\comment{
Two identical pinhole cameras are placed next to each other in a standard stereo camera setup. Let us denote the focal length with $f$, the horizontal and vertical pixel density with $\ve{k}=[k_h,k_v]$ (the number of pixels on the sensor for each world space unit of length), the baseline with $b$ and the principal point with $\ve{c}=[c_x,c_y]$. Following the calculations in \cite{barath2016novel} we can derive the affine matrix in this case. The general formula for the matrix that transforms the patch on image 1 to the corresponding patch on image 2 is the following \cite{barath2016novel}:

where $\nv=[n_x,n_y,n_z]$ denotes the surface normal and $\ve{p}=[x,y,z]$ denote the world space coordinates of the pixel.
}

Suppose we know the affine parameters $a_1$ and $a_2$. This yields two equations from Eq.\eqref{eq:affine_mtx}:
\begin{align}
    a_1(n_xx+n_yy+n_zz) =& n_x(x-b)+n_yy+n_zz,  
    \label{eq:a1} \\
    a_2f_y(n_xx+n_yy+n_zz) =& -bf_xn_y
    \label{eq:a2}
\end{align}
Let us introduce the following vectors if $\matr X=[x,y,z]^T$:
\begin{align}
    \ve{v_1} &= \matr X & 
    \ve{v_2} &= [x-b,y,z], \\
    \ve{v_3} &= f_y \matr X &
    \ve{v_4} &= [0,-bf_x,0]. 
\end{align}

\noindent If they are substituted to Eqs.\eqref{eq:a1} and \eqref{eq:a2}, after elementary modifications, then two new formulas can be written as:
\begin{equation*}
\begin{array}{cc}
    \dotp{\nv}{(a_1\ve{v_1}-\ve{v_2})}=0, &
    \dotp{\nv}{(a_2\ve{v_3}-\ve{v_4})}=0. 
\end{array}    
    \end{equation*}

\noindent Since $\nv$ is perpendicular to both $(a_1\ve{v_1}-\ve{v_2})$ and $(a_2\ve{v_3}-\ve{v_4})$, one can calculate $\nv$ up to a scalar multiple by computing their cross product. Expressed with the original parameters, this results in the following formula:
\begin{equation}
    \nv = \begin{bmatrix}
        k_hz(a_1-1)\\
        a_2k_vz\\
        -a_1k_hx-a_2k_vy+bk_h+k_hx
    \end{bmatrix}.
\end{equation}

\noindent where $k_h$ and $k_v$ denote the horizontal and vertical pixel densities of the camera sensor. Alternatively $f_x$ may be used instead of $k_h$ and $f_y$ instead of $k_v$.

\subsection{Calculating the Affine Parameters}
In order to use the formula presented above, we need to know the affine transformation between two corresponding image patches. Here, we present a novel robust method for the efficient calculation of the required two affine parameters from disparity maps \cite{matching_survey_hamzah2016literature}.

\begin{figure}
    \centering
    \includegraphics[width=0.6\linewidth]{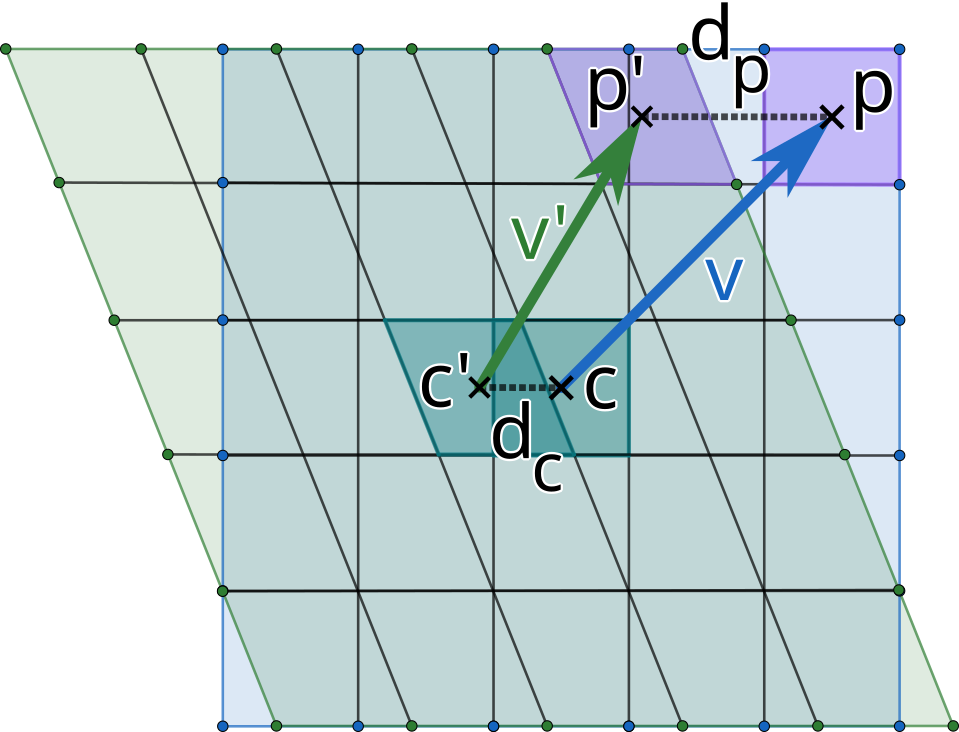}
    \caption{Two corresponding regions (squares) of the stereo image pair. The region in the left image is estimated by applying an affine transformation to its corresponding pair in the right image. This figure shows the original and the estimated regions overlapped. ($\mathbf c,\mathbf c'$) and ($\mathbf p,\mathbf p'$) denote two correspondences between the two regions with respective disparities $d_c$ and $d_p$. $\mathbf c,\mathbf c'$ are central pixels of the two regions, the pixels at where the affine transformation is estimated and normal calculated. $\mathbf v=\mathbf p-\mathbf c$, $\mathbf v'=\mathbf p'-\mathbf c'$. Note that  $\mathbf v'=\Aff \mathbf v$.}
    \label{fig:effine_equations}
\end{figure}

Let us examine a pair of corresponding points on the two images, and their surrounding pixels.
One of the patches transformed by the respective affine matrix approximately matches the corresponding area on the other camera image. The matching problem is visualized in \cref{fig:effine_equations}. If the disparity map is known, the following equation can be written:
\begin{equation}
    \vv + [d_p,0]^T = [d_c,0]^T + \vpv.
\end{equation}
Note that $d_p$ and $d_c$ can be negative when going from the left image to the right one.

Furthermore, $\vpv$ may be expressed using the affine transformation ($\vpv = \Aff \vv$), which yields
\begin{equation}
    \begin{bmatrix}
        d_p - d_c\\
        0
    \end{bmatrix} = 
    (\Aff -\mathbf I)\vv = 
    \begin{bmatrix}
        a_1-1 & a_2 \\
        0 & 0
    \end{bmatrix}
    \begin{bmatrix}
        v_x \\ v_y
    \end{bmatrix}.
\end{equation}

\noindent The upper row gives a linear equation for each surrounding pixel. With $N$ selected pixels, one can get an over-determined linear system of equations as follows:
\begin{equation}
    \underbrace{
    \begin{bmatrix}
        v_{1x} & v_{1y} \\
        v_{2x} & v_{2y} \\
        \vdots & \vdots \\
        v_{Nx} & v_{Ny}
    \end{bmatrix}}_{\matr V} \underbrace{\begin{bmatrix}
        a_1-1 \\
        a_2
    \end{bmatrix}}_\ve{a} = \underbrace{\begin{bmatrix}
        d_1 - d_c \\
        d_2 - d_c \\
        \vdots \\
        d_N - d_c
    \end{bmatrix}}_{\ve{d}}.
\end{equation}

\noindent Using the well-known least squares estimation of an inhomogeneous linear system of equations, we get the affine parameters as
    $\ve{a} = \left(\matr V^T \matr V\right)^{-1} \matr V^T\ve{d}$.

\subsection{Affine Parameter Calculation by Convolution}
Suppose the relative position of selected surrounding pixels is the same for all observed pixels. For example we always select an $n\times n$ square centered at the observed pixel. In this case, the entire $\matr S=\left(\matr V^T \matr V\right)^{-1}{\matr V}^T$ matrix can be precalculated as it only depends on the shape of the selected area, its disparity values do not influence the elements in $\matr S$. \textit{Precalculation of $\matr S$ can significantly reduce the time demand of the algorithm}. 

By expanding $\matr S$, the following formula is obtained for the convolution kernels:
\begin{equation}
    \matr V^T \matr V = \begin{bmatrix}
        \underbrace{\sum_{i=1}^N v_{ix}^2}_\alpha & 
        \underbrace{\sum_{i=1}^N v_{ix}v_{iy}}_\beta \\
        \underbrace{\sum_{i=1}^N v_{ix}v_{iy}}_\beta &
        \underbrace{\sum_{i=1}^N v_{iy}^2}_\gamma
    \end{bmatrix}.
\end{equation}
therefore:
\begin{eqnarray}
    \begin{split}
        &\matr S=\frac{1}{\alpha\gamma-\beta^2}\begin{bmatrix}
            \gamma & -\beta \\
            -\beta & \alpha
        \end{bmatrix}\begin{bmatrix}
        v_{1x} & \hdots & v_{Nx} \\
        v_{1y} & \hdots & v_{Ny} 
    \end{bmatrix} = \\
    &\frac{1}{\alpha\gamma-\beta^2}
    \begin{bmatrix}
        \gamma v_{1x} - \beta v_{1y} &
        \hdots &
        \gamma v_{Nx} - \beta v_{Ny} \\
        -\beta v_{1x} + \alpha v_{1y} &
        \hdots &
        -\beta v_{Nx} + \alpha v_{Ny}
    \end{bmatrix},
    \end{split}
\end{eqnarray}

When calculating the affine parameters, we just have to construct $\ve{d}$ for each observed pixel, then it multiplies by the precalculated elements of $\matr S$ from the left as
\begin{align}
\label{eq:convprep}
    a_1-1 &= \sum_{i=1}^N \underbrace{\frac{1}{\alpha\gamma-\beta^2}(\gamma v_{ix}-\beta v_{iy})}_{s_{1i}} \underbrace{(d_i-d_c)}_{\ve{d}_i}, \\
    \label{eq:convprep2}
    a_2 &= \sum_{i=1}^N \underbrace{\frac{1}{\alpha\gamma-\beta^2}(-\beta v_{ix}+\alpha v_{iy})}_{s_{2i}} \underbrace{(d_i-d_c)}_{\ve{d}_i}.
\end{align}

Note that other than the precalculated $\alpha,\beta$ and $\gamma$ constants, $s_{1i}$ and $s_{2i}$ are both only dependent on the relative position vector $\ve{v_i}$ of the $i$th pixel.

With some rearrangement, we can decompose the calculation of each parameter into two steps. A discrete 2D convolution of the disparity map with a special precalculated kernel, and a second step consisting of a multiplication by a precalculated constant and a subtraction:
\begin{align}
\label{eq:conv}
        a_1 & =\sum_{i=1}^Ns_{1i}(d_i-d_c)=\underbrace{\sum_{i=1}^Ns_{1i}d_i}_\textnormal{Step 1}\underbrace{-d_c\overbrace{\sum_{i=1}^Ns_{1i}}^{\delta_1}+1}_\textnormal{Step 2}
        , \\
        \nonumber
        a_2 &= \sum_{i=1}^Ns_{2i}(d_i-d_c) = \underbrace{\sum_{i=1}^Ns_{2i}d_i}_\textnormal{Step 1}\underbrace{-d_c\overbrace{\sum_{i=1}^Ns_{2i}}^{\delta_2}}_\textnormal{Step 2} .
\end{align}

 Inside the summations of \cref{eq:conv}, $s_{1i}$ and $s_{2i}$ are only multiplied with the disparity of the pixel at $\ve{v_i}$. Therefore, this is a 2D convolution with the kernel containing $s_{1,1},s_{1,2},\ldots,s_{1,N}$ for the computation of $a_1$, and $s_{2,1},s_{2,2},\ldots,s_{2,N}$ for $a_2$.

\begin{figure}
    \centering
    \begin{tikzpicture}
\draw[step=1cm,color=gray] (-2,-2) grid (1,1);

\newcounter{mycountrow}
\newcounter{mycountcol}
\newcounter{mycounttotal}
\setcounter{mycountrow}{-1}
\setcounter{mycountcol}{-1}
\setcounter{mycounttotal}{1}
\foreach \y in {+0.75,-0.25,-1.25} {
  \setcounter{mycountcol}{-1}
  \foreach \x in {-1.5,-0.5,0.5} {
    \newcommand{\asd}{\node at (\x,\y) {$s_{1, \arabic{mycounttotal}}$};}
    \newcommand{\asdb}{\node at (\x,\y-0.5) {$\scriptstyle\left[\arabic{mycountrow}, \arabic{mycountcol}\right]$};}
    
        \asd
        \asdb
    \addtocounter{mycounttotal}{1}
    \addtocounter{mycountcol}{1}
    }
    \addtocounter{mycountrow}{1}
    }
\end{tikzpicture}
    \caption{An example of a $3\times 3$ kernel for computing $a_1$. The listed coefficients $s_i$ and their corresponding vectors $\ve{v_i}^T$ are listed in the same cell.}
    \label{fig:example-kernel3x3}
\end{figure}

In both cases, the first step is a 2D convolution of the disparity map with two kernels. This is followed by the other operations involving the precomputed $\delta_1$ and $\delta_2$, defined in \cref{eq:conv}. An example kernel layout is shown in  \cref{fig:example-kernel3x3}.

The bulk of the computation is in the calculation of the 2D convolutions. In practice, this is efficiently sped up using the Fast Fourier Transform (FFT) algorithm either on CPU or GPU \cite{fft, fftgpu}. However, according to our testing, on modern GPUs and with reasonable kernel sizes, the method is already very fast without the additional performance gain of FFT.

\subsection{Adaptive Region Estimation}
A very important challenge for robust surface normal estimators is maintaining accuracy around corners and edges. The main problem here is that including points from a neighboring surface will yield less accurate normals. This usually appears as a smoothing of normals around edges, which is especially problematic when the results are used for segmentation tasks. Nan Ming \etal~\cite{sda-sne} propose minimizing the depth Laplacian via dynamic programming. Yi Feng \etal~\cite{d2nt} also propose a post processing step for correcting the smoothing effect around edges.

We introduce a dynamically shaped kernel for our method described above. The main idea is to iteratively extend the patch of included points, starting from the observed pixel. 

\begin{figure}
    \centering
    \includegraphics[width=0.45\linewidth]{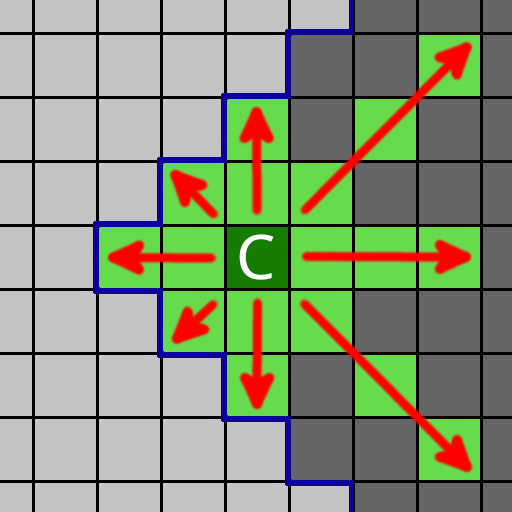}
    \caption{Visualization of the selected (green) pixels by performing star-fill from C in eight directions.}
    \label{fig:star-fill}
\end{figure}

A natural method of choice would be to use flood filling for the selection of included pixels. However, in our case, we have to run this selection algorithm for each pixel. Instead we introduce a simplified technique, for the sake of a more efficient GPU implementation. Our algorithm starts from the observed pixel at the center and traverses several lines in $M$ uniformly distributed directions. The traversal either stops when a maximum length is reached, or when a stopping condition is met. All visited pixels are included in the final normal estimation. The selected pixels will be in star-like shape, visualized in \cref{fig:star-fill}. More details on star filling and the considerations for choosing this method are given in the supplementary material.

Once we have a list of included pixels, we need to do two passes.
During the first pass, parameters $\alpha,\beta$ and $\gamma$ are computed. In the second pass, the affine parameters are estimated using \cref{eq:convprep}. and \cref{eq:convprep2}.

\noindent\textbf{Stopping at edges on the depthmap. (Simple thresholding, ST)}
As a first step, an edge detection filter of choice is run over the depth map. It stops when an edge pixel surpassing a threshold is reached. We use the depth Laplacian as a measure of discontinuity.

\noindent\textbf{Stopping based on covered depth. (CD)}
Let us assume that the central pixel has depth $d$, the surrounding pixels on the same flat surface should be in a range of $kd$, where $k$ is an adaptive threshold parameter.

With this method, we keep track of the maximum and minimum encountered depth along our traversal. If their difference surpasses $kd$, then the traversal is stopped. This method adapts much better to wider ranges. Unfortunately, it usually misses depth-continuous edges, like the edge detection based method.

More on stopping conditions are included in the supplementary material.


\section{Experimental Results}
The dataset used for evaluation is the 3F2N dataset \cite{fan2021three}, which contains 24 synthetic images with ground truth normals and depth values. We calculated the disparity maps from the depth values, with a unit length baseline (1 meter). Then we added noise with a Gaussian distribution to generate our tests. The dataset is divided into three difficulty levels (Easy, Medium, Hard). Scenes with a higher difficulty contain more discontinuities and fine detail.

The proposed method has been implemented in an OpenGL shader, along with some other methods for reference. Tests have been conducted on a laptop equipped with an Nvidia RTX 4060 (mobile) GPU and an AMD Ryzen 7 7840HS CPU. 

\noindent \textbf{Implemented algorithms.} We have developed all variants of the proposed approach. Additionally, a PCA-based solution and a naive implementation are also examined in the tests. The naive solution is denoted by 'Cross' in the tests, when the normal is calculated from two tangent vectors. These vectors are determined by triangulation of the processed point and the horizontal and vertical neighbouring pixels.

The PCA-based normal estimation is also our own implementation, we set its parameters by an exhaustive search to tune it for the best results. The essence of the estimation is the eigenvalue/eigenvector computation of a $3 \times 3$ matrix, we applied closed-form formulas for the sake of speed.

Moreover, we tested the SDA-SNE \cite{sda-sne} method as well, which is claimed to be robust to noise. Unfortunately, a GPU implementation for the technique is not available at the moment, therefore we compared with the original CPU implementation, mainly for comparing accuracy.

\begin{figure}
    \centering
    \includegraphics[width=1\linewidth]{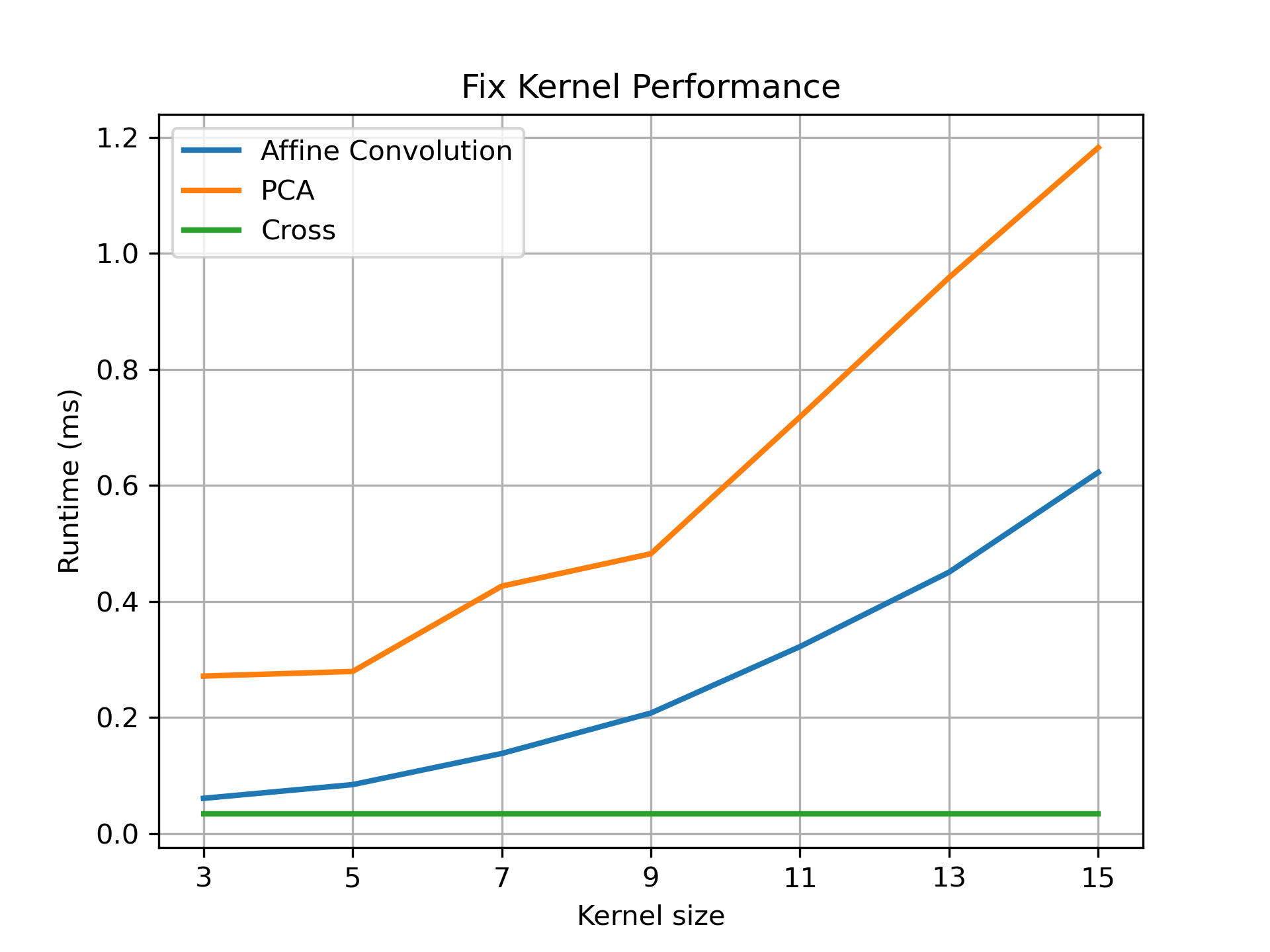}
    \caption{Comparison of execution time depending on kernel size. The naive cross product execution time is also displayed for reference. Average over 100 frames at resolution 480x620.}
    \label{fig:perf-kernel}
\end{figure}

\subsection{Runtime}
To avoid overhead from unrelated rendering tasks, we exclusively measured the GPU execution time of the shaders. This was accomplished utilizing OpenGL query objects to get the elapsed GPU time. A comparison between the adaptive methods, PCA, and the fixed kernel version is shown in \cref{tab:perf-disc} and \cref{tab:perf-disc2}. 

The provided SDA-SNE implementation \cite{sda_code} is for the CPU. To compensate for at least the additional time taken by the python interpreter, we ran the original python script through cython, compiled it and ran all tests natively.

Additionally, we have compared the execution time of the methods depending on the kernel size. The affine convolution method shows about twice the performance of PCA. Time demands are reported in~\cref{fig:perf-kernel}.

It is clearly seen that adaptive methods are slower, as it is expected, but all the values are below $10 ms$, therefore real-time operation can be reached even if large kernel size or adaptive disparity selection are applied.

\sisetup{detect-weight,
round-mode = places,
round-precision = 2
}

\begin{table}
  \centering
  \begin{tabular}{ l*{4}{S[table-format=2.1]}}
    \toprule
    {Method} & {Avg (ms)} & {Min (ms)} & {Max(ms)}  & {Std (ms)} \\
    \midrule
Cross&0.03&0.03&0.03&0.00\\
PCA 9x9&0.48&0.48&0.48&0.00\\
\textbf{Affine 9x9}&0.21&0.21&0.21&0.00\\
\textbf{ST s10 d8 t0.1}&0.19&0.12&0.30&0.04\\
\textbf{CD s10 d8 t0.5}&0.24&0.17&0.50&0.07\\
SDA-SNE&190.35&162.48&220.53&15.25\\
    \bottomrule
  \end{tabular}
  \caption{Statistics of frametimes over 100 frames at a resolution of 480x640. 's' denotes the maximal number of steps taken in each direction, 'd' denotes the number of uniformly distributed directions. The threshold for both conditions is denoted with 't'. All measurements are in milliseconds.}
  \label{tab:perf-disc}
\end{table}

\sisetup{detect-weight,
round-mode = places,
round-precision = 2
}
\begin{table}
  \centering
  \begin{tabular}{ l*{5}{S[table-format=2.1]}}
    \toprule
    {Method} & {Avg} & {Min} & {Max} & {Med}  & {Std} \\
    \midrule
    {Affine 9x9} & \B 0.64965152 & \B 0.61712 & \B 0.71856 & \B 0.651536 & \B 0.014497420712719903\\
PCA 9x9 & 1.6216513599999998 & 1.548832 & 1.734112 & 1.617616 & 0.029585101599392895\\
{ST[s:10 d:8]} & 1.795310208 & 1.7272 & 2.070144 & 1.793424 & 0.030285530656680524\\
{ST[s:30 d:16]} & 8.146468832 & 7.728896 & 8.891392 & 8.14232 & 0.2107553613535745\\
{CD[s:10 d:8]} & 1.712877344 & 1.664416 & 1.861568 & 1.70704 & 0.0308639819696951\\
{CD[s:30 d:16]} & 6.853837088000001 & 6.466208 & 8.009472 & 6.844608 & 0.20761573227533664\\
    \bottomrule
  \end{tabular}
  \caption{Statistics of frametimes over 1000 frames at a resolution of 1024x720. Both of the adaptive methods were measured with 8 directions and max 10 steps as well as 16 directions and max 30 steps. The threshold for both conditions was set to 0.1. All measurements are in milliseconds.}
  \label{tab:perf-disc2}
\end{table}

\subsection{Robustness to Noise}
To measure the effect of disparity noise at different viewing angles, we rendered a sphere placed directly in front of the camera. For better accuracy, the sphere has been rendered via raycasting \cite{roth1982ray}, the true normals were calculated per pixel. 
Disparities were calculated by projecting the ray-sphere intersection to two virtual pinhole cameras and taking the difference of the projected coordinates.
Their noise is assumed to have a Gaussian distribution with zero mean. Therefore, we add this $\mathcal{N}(0,\sigma)$ noise to the calculated disparities.

\begin{table}
    \centering
    \title{Estimation errors on a sphere with disparity noise }
    \begin{tabular}{ l*{2}{S[table-format=2.1]}}
         \toprule
        {Method} & {$\sigma=0.2$} & {$\sigma=1$} \\
        \midrule
        Affine 3x3 & 19.153201029603828 & 51.743688361677755\\
Affine 5x5 & 6.91897812804246 & 30.523083501870886\\
Affine 9x9 & 2.2150090460352554 & 10.471874644887858\\
Affine 15x15 & \B 0.9717758224861539 & \B 3.9366604242212273\\
PCA 3x3 & 27.442218763619838 & 50.357805560212114 \\
PCA 5x5 & 11.218555434474 & 42.85188095102227\\
PCA 9x9 & 3.6216701958802315 & 35.299802013122175\\
PCA 15x15 & 1.554688882962494 & 24.4180143444789\\
         \bottomrule
    \end{tabular}
    \caption{Average angle in degrees between ground truth and estimated normal vectors for a sphere with $\mathcal{N}(0,0.2)$ and $\mathcal{N}(0,1)$ disparity noise (Resolution: 1024x1024). Note that with the same kernel size the affine-based method performs significantly better.}
    \label{tab:noise-compare}
\end{table}

\cref{tab:noise-compare} shows our results for different kernel sizes and for a scenario with $\sigma=1$ px and a more reasonable case with $\sigma=0.2$ px. For reference, the Middlebury dataset's "Adirondack" scene has an average standard deviation of $0.12$ over the samples. In our measurements, the affine convolutional estimator exhibited superior accuracy and consistency over PCA, especially for larger kernel sizes.

\begin{figure}
\centering
\begin{subfigure}{0.48\columnwidth}
    \includegraphics[width=\textwidth]{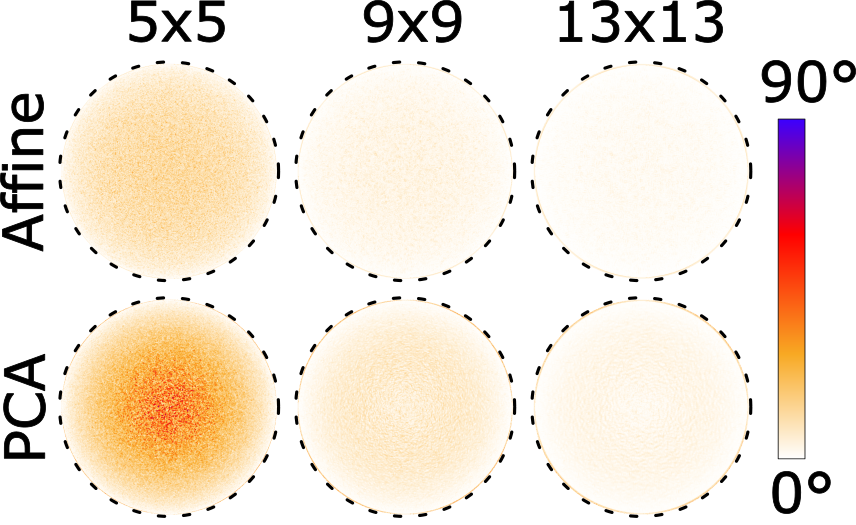}
    \caption{$\mathcal{N}(0,0.25)$}
    \label{fig:sphere-n02}
\end{subfigure}
\hfill
\begin{subfigure}{0.48\columnwidth}
    \includegraphics[width=\textwidth]{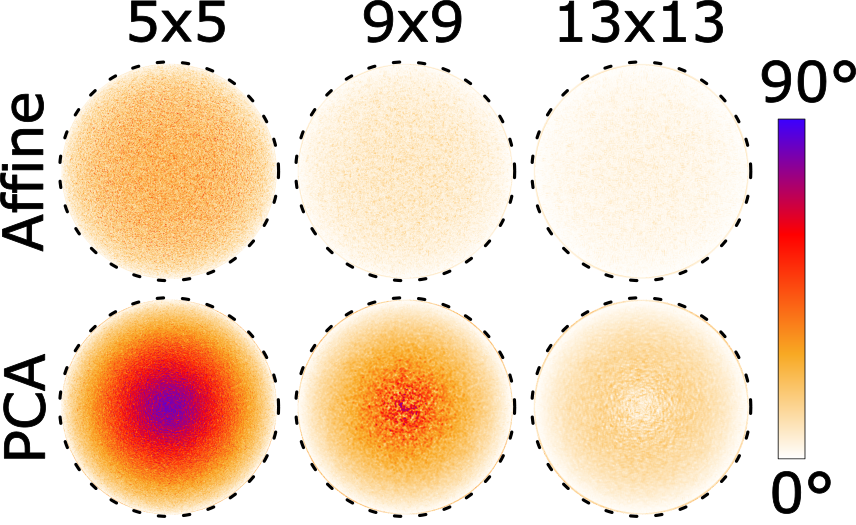}
    \caption{$\mathcal{N}(0,0.5)$}
    \label{fig:sphere-n05}
\end{subfigure}
\begin{subfigure}{0.48\columnwidth}
    \includegraphics[width=\textwidth]{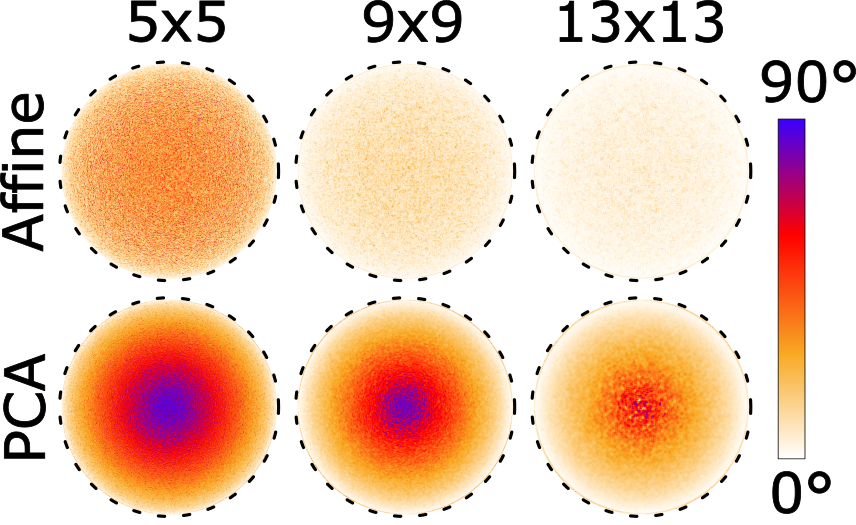}
    \caption{$\mathcal{N}(0,0.75)$}
    \label{fig:sphere-n075}
\end{subfigure}
\hfill
\begin{subfigure}{0.48\columnwidth}
    \includegraphics[width=\textwidth]{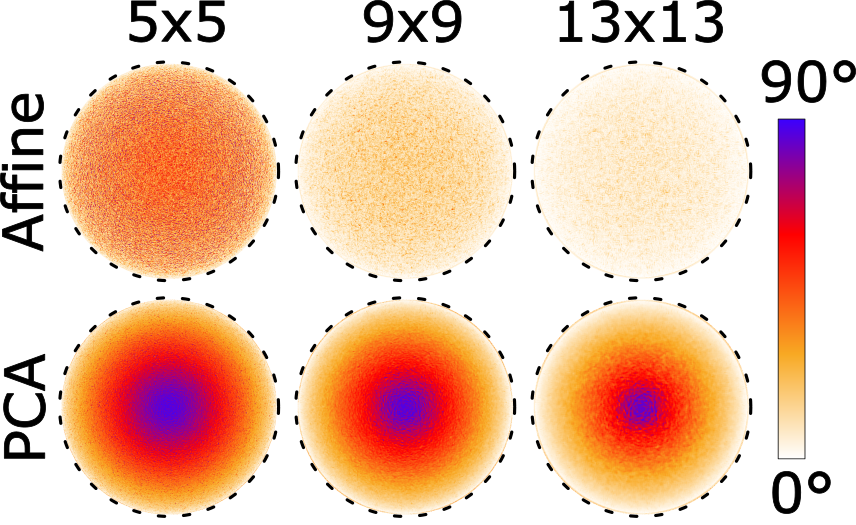}
    \caption{$\mathcal{N}(0,1)$}
    \label{fig:sphere-n1}
\end{subfigure}
       
\caption{Angular normal error of a sphere. Radius: 1.4, view distance: 3 units from center, baseline: 0.3, resolution: 1024x1024.}
\label{fig:figures}
\end{figure}

\begin{figure}
    \centering
    \includegraphics[width=0.8\linewidth]{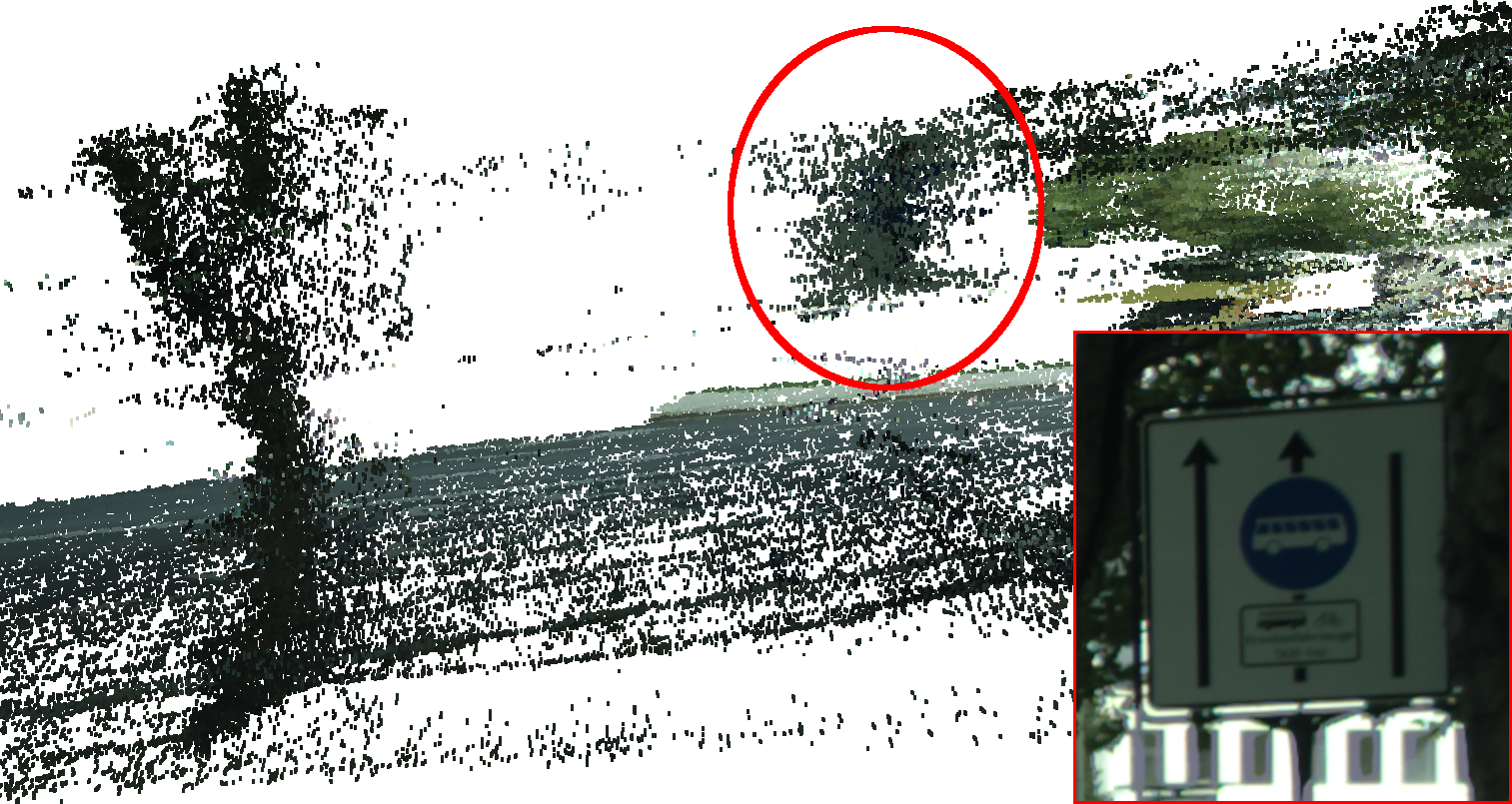}
    \caption{Side view of pointcloud generated from disparities (Cityscapes dataset). The points of the sign in the bottom right corner are circled in the center. As  errors affect the depth of the points more, the pointcloud is distorted here with a very high signal to noise ratio. }
    \label{fig:sign}
\end{figure}

As shown in \cref{fig:sphere-n02} and \cref{fig:sphere-n1}, the sensitivity to noise tends to depend on the viewing angle. The reasons of it is that the noise in the disparities does not affect all directions uniformly, yielding a distortion to the general shape of the point-cloud. An example for this effect is shown in \cref{fig:sign}. 
We think that this non-uniform scaling is the reason behind PCA performing worse especially in heavy noise circumstances. Therefore, \textit{the use of affine transformations has theoretical advantages over PCA-based solution(s).}

\begin{figure}
    \centering
    \includegraphics[width=0.9\linewidth]{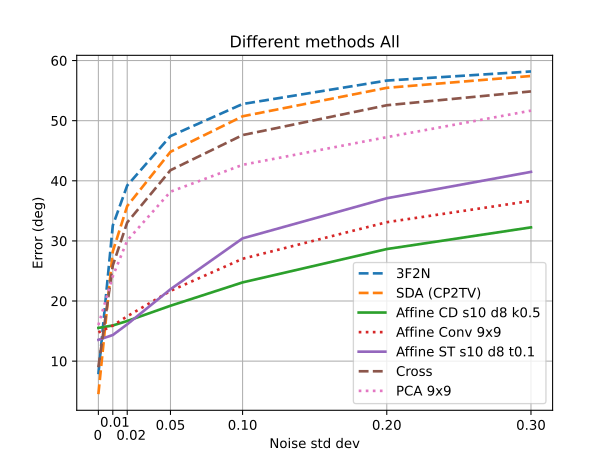}
    \caption{Comparison of average angular error depending on disparity noise. Complete dataset, 
 images of 480x640 resolution.}
    \label{fig:all-methods}
\end{figure}

Additionally, Fig. \ref{fig:all-methods} shows the accuracy of all methods depending on noise level. SDA-SNE did not show any noise filtering capability, opposed to the claim in the original paper \cite{sda-sne}.

%

\begin{figure*}
    \centering
    \begin{subfigure}{0.04\linewidth}
    \centering
    \includegraphics[height=2.3cm]{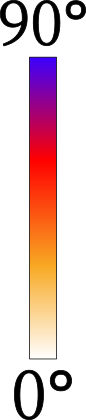}
    \vspace{0.2cm}
    \end{subfigure}
    \begin{subfigure}{0.23\linewidth}
    \centering
    \includegraphics[width=\linewidth]{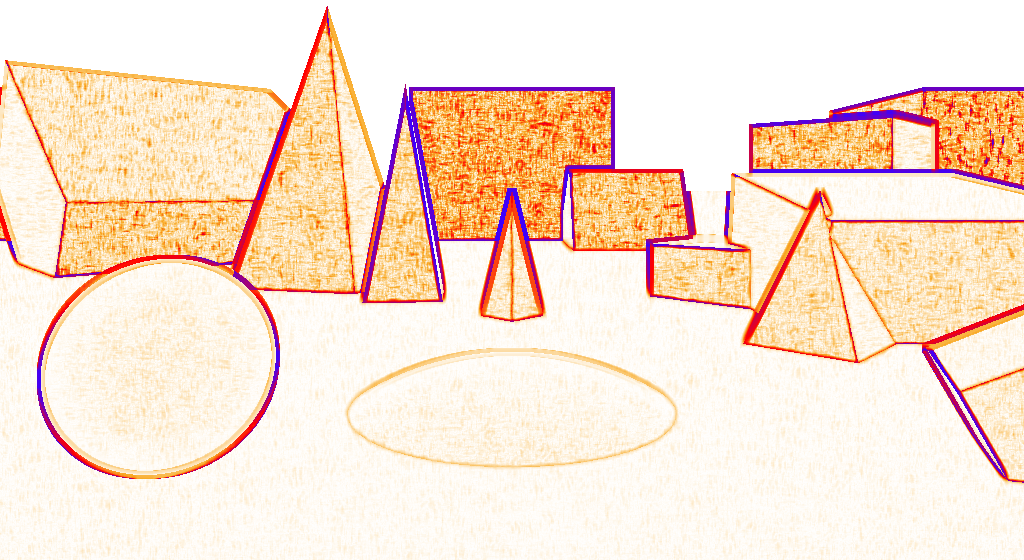}
    \caption{Affine 9x9}
    \label{fig:scene-lsq}
    \end{subfigure}
    \begin{subfigure}{0.23\linewidth}
    \centering
    \includegraphics[width=\linewidth]{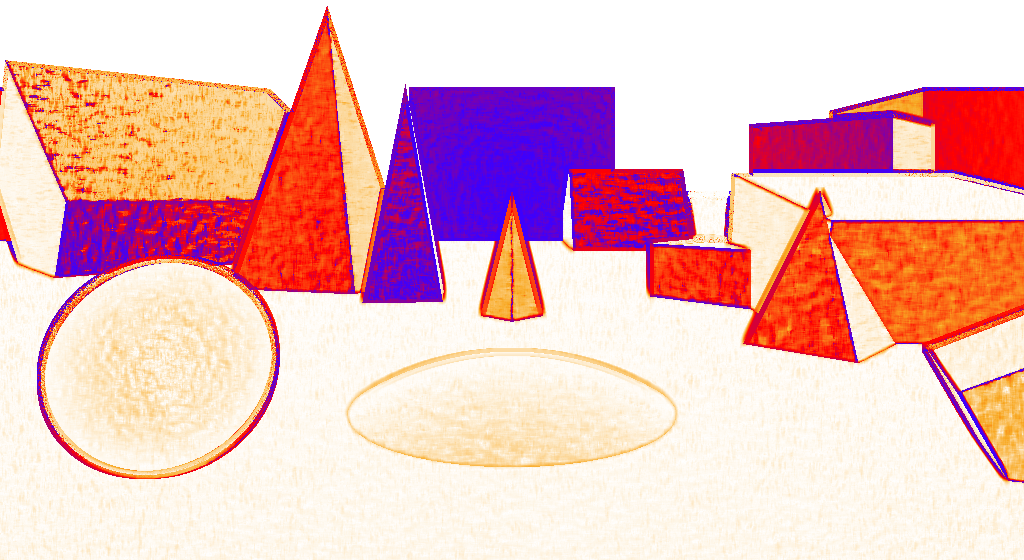}
    \caption{PCA 9x9}
    \label{fig:scene-pca}
    \end{subfigure}
    \begin{subfigure}{0.23\linewidth}
    \centering
    \includegraphics[width=\linewidth]{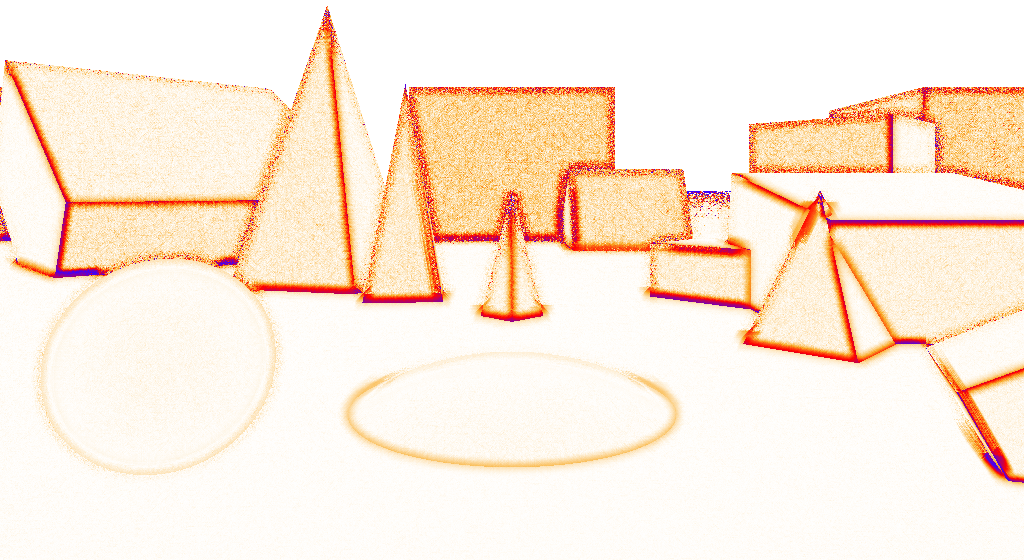}
    \caption{CD s:10 d:8}
    \label{fig:scene-cd-s10-s8}
    \end{subfigure}
    \begin{subfigure}{0.23\linewidth}
    \centering
    \includegraphics[width=\linewidth]{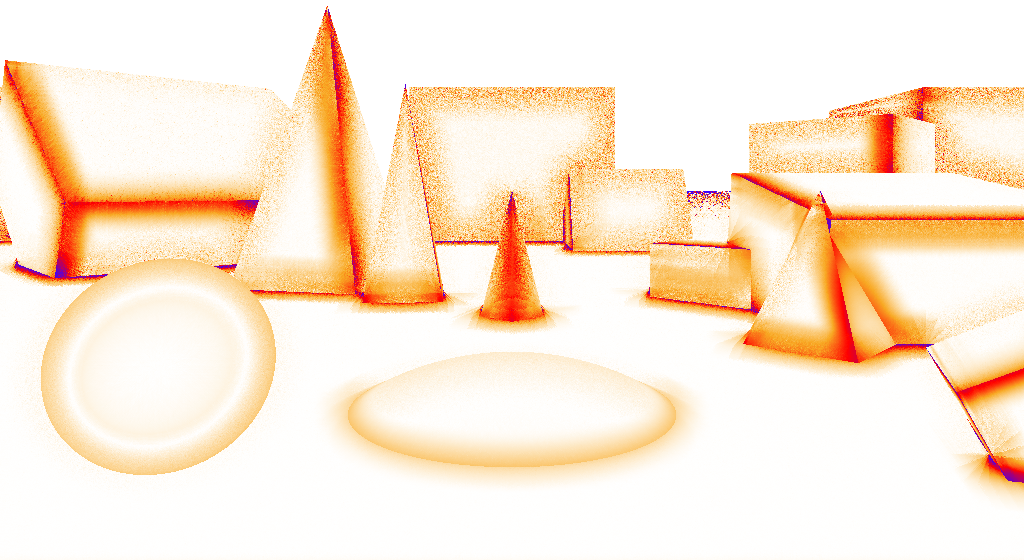}
    \caption{CD s:30 d:16}
    \label{fig:scene-cd-s30-s16}
    \end{subfigure}
    \caption{Angular error of estimated normals on synthesized scene. Angles in degrees. $\mathcal{N}(0,0.2)$ noise. The furthest box in the scene is of size $4\times3\times4$ and its center is 15.5 units from the camera focal point. baseline: 0.3, resolution: 1024x720.}
    \label{fig:scene}
\end{figure*}

\paragraph{Discontinuities and the adaptive version.}
\Cref{fig:scene} shows a comparison of the discussed methods on a test scene, consisting of multiple simple shapes. On this scene, the furthest cube is 15 units away from the cameras. The used baseline is 0.3 and the applied noise is of the distribution $\mathcal{N}(0,0.2)$. \cref{tab:scene-errors} contains the numerical measurements excluding background pixels. The runtime performance of calculating the normals for this scene is displayed in \cref{tab:perf-disc2}. 

For the same amount of included pixels, the adaptive version is much slower than the convolutional method. Using star fill with $8$ directions and maximum $10$ steps takes about thrice the amount of time it takes to run the convolutional method with a 9x9 kernel, while processing a similar number of pixels. Despite being slower, star fill still came close to the performance of 9x9 PCA in this case, with much better accuracy.

Most of the errors come from areas close to edges. There are two reasons for this. There are cases when the stopping condition is not triggered. This occurs mostly around edges where the depth is exactly $C^0$ continuous. In other cases, the stopping condition is rightly triggered, reducing the number of included points, but there are not enough remaining points for an accurate estimation. This is usually observable around strong discontinuities, such as around the edges of the furthest cube in the center of ~\cref{fig:scene-pca}.

\begin{table}
    \centering
    \title{Estimation errors corresponding to \cref{fig:scene}}
    \begin{tabular}{ l*{4}{S[table-format=2.1]}}
        \toprule
        {Method} & {Avg} & {Min} & {Max}  & {Std} \\
        \midrule
        Affine 9x9 & 9.79058192998059 & \B 0.0 & \B 89.994131325  & 16.26573457203153\\
PCA 9x9 & 22.068400319700032 & 0.019799999999999818 & 89.995901589 & 28.067565439484817\\
ST[s:10 d:8] & 8.272856388359559 & \B 0.0 & \B 89.99141157 & 13.802423055536341\\
ST[s:30 d:16] & \B 8.075060141835353 & \B 0.0 & 89.996126886  & 15.224192565815404\\
CD[s:10 d:8] & 6.621080568791684 & \B 0.0 & 89.996126886 & \B 12.109603037407725\\
CD[s:30 d:16] & 7.523658087615706 & \B 0.0 & 89.996126886 & 12.126347875105523\\
         \bottomrule
    \end{tabular}
    \caption{Angle in degrees between ground truth and estimated normal vectors with $\mathcal{N}(0,0.2)$ disparity noise. (Scene shown on Fig. \ref{fig:scene})}
    \label{tab:scene-errors}
\end{table} 

\begin{figure}
    \centering
    \includegraphics[width=0.8\linewidth]{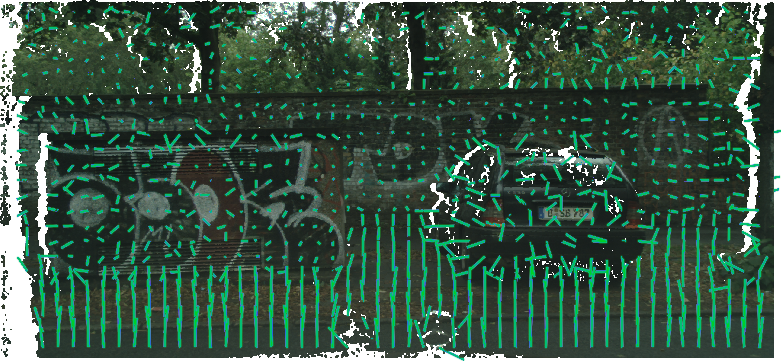}
    \includegraphics[width=0.8\linewidth]{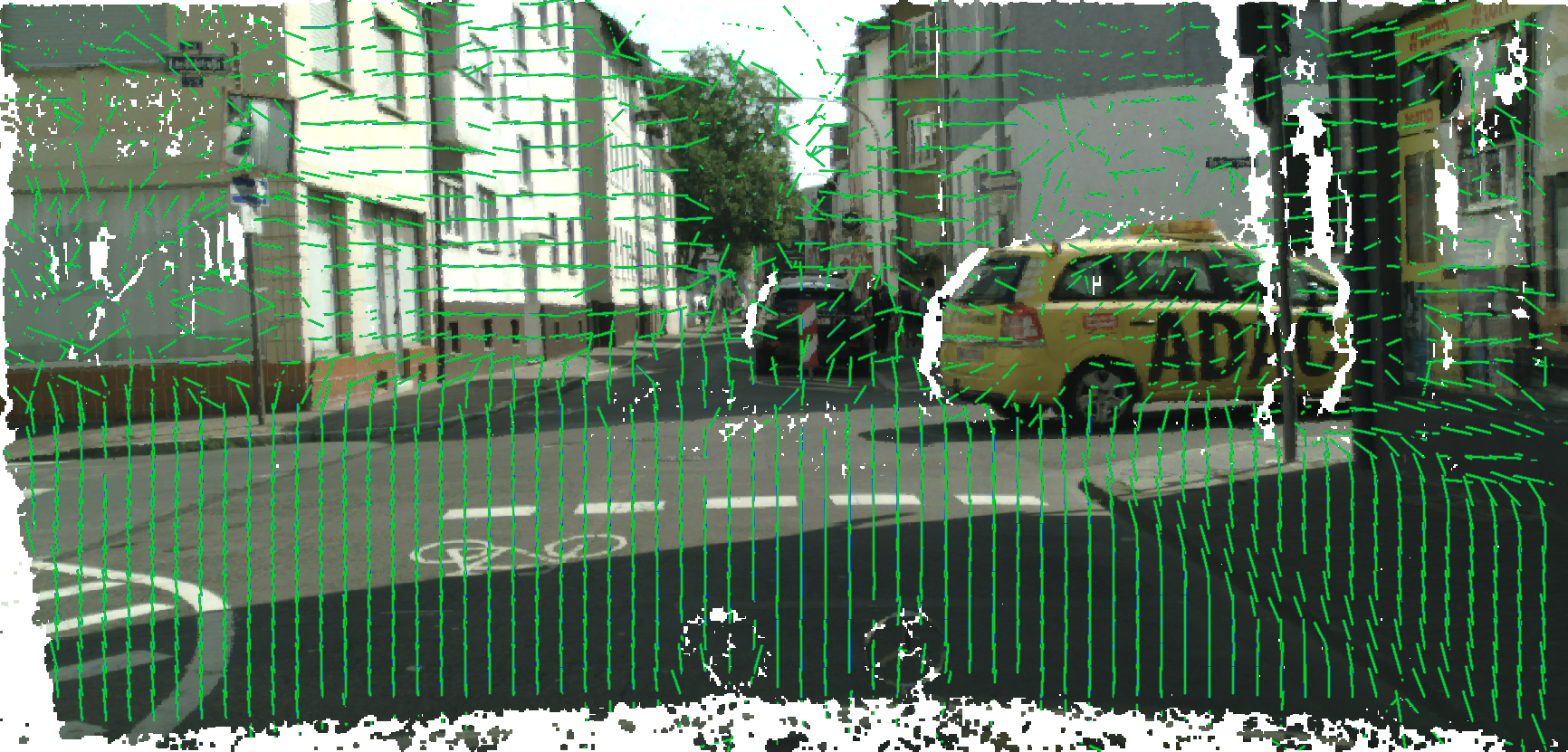}
     \includegraphics[width=0.8\linewidth]{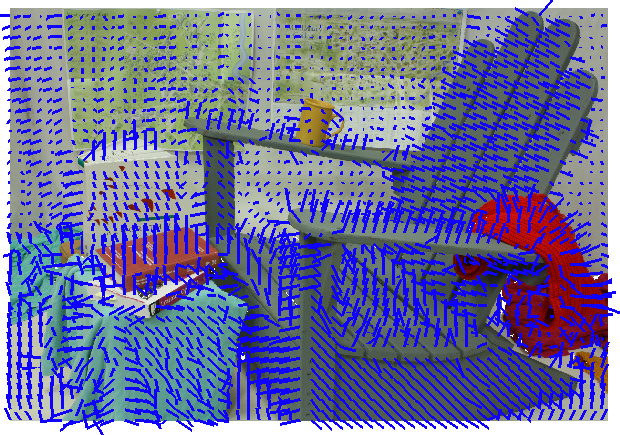}
    \caption{Some real-world examples with estimated normals from the Cityscapes \cite{Cordts2016Cityscapes} (above) and the Middlebury \cite{scharstein2014high} dataset (below).}
    \label{fig:realworld}
\end{figure}

Some images of real world results are provided on \cref{fig:realworld}.
Numeric stability tests are provided in the supplementary.

\section{Conclusions}

It is shown here how affine correspondences provide an effective way of estimating surface normals from rectified stereo images. GPU implementations guarantee efficient and rapid estimation. Our solution is purely geometric, without any machine learning-based component. 

For fixed sized and shaped kernels, the proposed method outperforms PCA-based normal estimation both in terms of runtime and accuracy. The experiments show double speed-up w.r.t. an optimized PCA implementation while offering about $30\%-40\%$ improvement over the average angular error. Moreover, we introduced heuristic methods for adapting our affine estimator to dynamic neighbour selection. They managed to improve accuracy on our test scene by $20-25\%$ over the fixed kernel estimator. 

Distant camera facing surfaces are challenging for normal estimation as these scenarios exhibit a very high signal to noise ratio. 
Affine methods are more efficient in filtering the noise in such extreme conditions. This is due to not relying directly on the geometry of the triangulated points.

In conclusion, a novel approach is presented here for robust and accurate surface normal estimation from rectified stereo images. Its GPU-powered implementation is able to run even in real-time systems. 

\section{Acknoweledgment}
This research has been supported by scholarship from Robert Bosch Kft.


    \bibliographystyle{IEEEtran}
    \bibliography{references.bib}

\end{document}